\def\ARXIVVERSION{1}

\documentclass[letterpaper,10pt,conference]{ieeeconf}

\IEEEoverridecommandlockouts

\usepackage{amsmath}
\usepackage{amssymb}
\usepackage{array}
\usepackage{booktabs}
\usepackage{xcolor}
\usepackage{cite}
\usepackage{graphicx}
\makeatletter
\ifx\endfigure\end@float\def\endfigure{\end@float}\fi
\ifx\endtable\end@float\def\endtable{\end@float}\fi
\makeatother
\usepackage[caption=false,subrefformat=subparens]{subfig}
\DeclareCaptionFont{mpepanel}{\fontsize{6.3}{7.5}\selectfont}
\usepackage{hyperref}

\ifdefined\ARXIVVERSION
\hypersetup{
  hidelinks,
  pdftitle={AeroWeaver: An Embodied-Agent Harness for Weaving Aerial Skills into Distributed, Adaptive Swarm Execution},
  pdfauthor={Jiabin Lou, Yirong Yang, Haopeng Wang, Xuxin Lv, Xinyu Liu, Diyuan Hou, Xuehong Liu, Rongye Shi, Wenjun Wu}
}
\else
\hypersetup{hidelinks}
\fi
\begin{document}
\bstctlcite{IEEEexample:BSTcontrol}

\title{\LARGE \bf
AeroWeaver: An Embodied-Agent Harness for Weaving Aerial Skills into Distributed, Adaptive Swarm Execution}

\ifdefined\ARXIVVERSION
\author{%
Jiabin Lou, Yirong Yang, Haopeng Wang, Xuxin Lv, Xinyu Liu,\\
Diyuan Hou, Xuehong Liu, Rongye Shi, and Wenjun Wu$^{*}$%
\thanks{All authors are with Beihang University, 37 Xueyuan Road, Haidian District, Beijing 100191, China.}%
\thanks{$^{*}$Corresponding author: Wenjun Wu (\texttt{wwj09315@buaa.edu.cn}).}%
}
\else
\author{Jiabin Lou, Yirong Yang, Haopeng Wang, Xuxin Lv, Xinyu Liu, Diyuan Hou, Xuehong Liu, Rongye Shi, and Wenjun Wu}
\fi

\maketitle

\begin{abstract}
Collective intelligence is a collaborative autonomy paradigm in which multiple agents pursue shared objectives through local perception, information exchange, and coordinated action. UAV swarms embody this paradigm by coordinating multiple vehicles in tasks such as search, inspection, and tracking. Recent advances in large language model (LLM) agents have strengthened natural-language task understanding and high-level planning, providing a flexible semantic interface between mission descriptions and collective behavior.
While these advances expand semantic reasoning, applying LLM agents to UAV swarms raises challenges in grounding model decisions in executable capabilities, reconciling global task reasoning with distributed execution, and using mission-specific experience for continual adaptation. To address these challenges, we introduce AeroWeaver, an embodied-agent harness that weaves individual UAV skills into coordinated mission-level behavior. AeroWeaver connects semantic decisions to governed skills, organizes role-conditioned local agents for distributed coordination, and uses role-indexed state–action–reward experience to refine skill selection online. Experiments and runtime validation show that AeroWeaver maintains valid skill execution under tested conditions and supports body-local multi-UAV operation without a central agent generating joint actions from global context, while reward-guided online updates provide a training-free path for adaptive learning swarm agents from accumulated execution experience. Code website:
\ifdefined\ARXIVVERSION
\url{https://github.com/Admire-ljb/AeroWeaver}.
\else
\url{https://github.com/Admire-ljb/AeroWeaver}.
\fi

\end{abstract}

\section{Introduction}
\label{sec:introduction}

Collective intelligence enables multiple agents to combine local perception, information exchange, and coordinated action around a shared objective.
UAV swarms provide a representative physical realization by organizing the sensing, mobility, and task capabilities of multiple aerial platforms.
As missions and operating conditions change, their collective effectiveness depends on a continuous link between high-level task organization and reliable vehicle-level execution.

Recent advances in large language models (LLMs) bring complex-task understanding, knowledge organization, hierarchical reasoning, and long-horizon planning to the task-organization layer.
These capabilities make LLMs a flexible interface for interpreting mission descriptions and structuring high-level swarm behavior~\cite{iannoli2026mission}.
The resulting decisions ultimately operate through vehicle-specific observations, communication links, and actuators, making their connection to distributed physical execution a central research problem.

AI-agent systems mediate this connection through the runtime surrounding the model, commonly termed a harness.
The harness supplies context, exposes actions, routes model selections to executable tools, and returns environmental feedback.
For an embodied swarm, the harness therefore becomes the interface between shared mission reasoning and multiple physically separate decision and execution processes.
Viewed across the full execution loop, this interface raises a connected sequence of questions concerning whether semantic choices are physically executable, how executable choices remain coordinated across distributed vehicles, and how their outcomes should influence later decisions.
The sequence begins at the semantic-to-physical boundary, where a language-level choice must correspond to a capability available in the deployed system and be routed to the vehicle that owns its execution interface.
Without this mapping, task reasoning and flight control remain separate software layers.

Even when each action is executable, swarm-scale coordination remains difficult because task-level reasoning and body-level execution follow different information topologies.
Many LLM-based multi-robot systems aggregate team state in a shared planner that decomposes the mission and returns a joint plan~\cite{li2026llmmrs}.
This organization places semantic decisions in a shared context, whereas observations, communication links, and actuators remain distributed across vehicles.
The coordination problem is to maintain coherent mission progress across this distributed execution topology.

As these distributed decisions accumulate across rounds, the execution loop produces experience that can inform later choices.
The relevance of each record depends on the local state, vehicle responsibility, peer interaction, and stage of mission progress in which it was collected.
The adaptation problem is to relate accumulated outcomes to the current decision while preserving stable physical execution.

Together, these requirements call for a swarm runtime that connects semantic capability selection, body-local coordination, and experience-based adaptation within a consistent execution boundary.
AeroWeaver addresses this need as an embodied-agent harness that weaves individual UAV skills into coordinated swarm execution.
Its skill interface connects mission interpretation to body-scoped capability dispatch, while task-conditioned contexts preserve local decision authority across vehicles and reward-linked experience adjusts subsequent skill preferences without changing model weights or low-level controllers.
The resulting loop allows mission-level reasoning to organize collective behavior, body-bound agents to execute through their assigned platforms, and accumulated outcomes to inform later decisions.

Our contributions are as follows:
\begin{itemize}
  \item \textbf{Embodied-Agent Harness with Aerial Skills.} We represent UAV capabilities as typed skill packages that pair \texttt{skill.md} documentation with executable objects and route selected skills through body-scoped executors.
  \item \textbf{Distributed Swarm Orchestration.} We derive role-specific prompts and active skill subsets from each mission, bind each local agent to a UAV, and incorporate directed peer messages into the next body-local decision without introducing a central action controller.
  \item \textbf{Experience-Guided Online Reinforcement.} We index state--action--reward trajectories by semantic role, estimate skill-level advantages from similar experience, and enable training-free online policy optimization through experience-guided score updates.
\end{itemize}

The evaluation examines task performance and inference cost in MPE-inspired simulation scenarios and traces how accumulated rewards influence subsequent skill choices.


\section{Related Work}
\label{sec:related-work}

\subsection{LLM-Based Embodied Agents}

LLM-based embodied agents connect natural-language reasoning with perception, planning, and action in interactive environments.
ReAct interleaves reasoning with actions and environmental observations, allowing plans to be revised during interaction~\cite{yao2023react}.
For physical robots, the representation linking this reasoning to control varies across systems.
SayCan combines language-model scores with learned affordance values to select feasible robot skills~\cite{saycan}.
Code as Policies generates programs that compose perception and control APIs~\cite{liang2023codepolicies}.
ReKep instead expresses manipulation goals as relational keypoint constraints and obtains actions through hierarchical optimization~\cite{huang2025rekep}.
Voyager accumulates executable programs during interaction and retrieves them as reusable skills for later tasks~\cite{Voyager}.
These approaches establish skills, programs, and geometric constraints as intermediate representations between language-level reasoning and executable behavior.

Repeated interaction also provides experience that can improve later decisions.
ExpeL extracts transferable natural-language knowledge from prior trajectories~\cite{zhao2024expel}.
Agent Workflow Memory identifies recurring workflows in agent trajectories and retrieves them during subsequent online or offline decisions~\cite{wang2025workflow}.
A-Mem organizes experience as linked notes whose contextual attributes evolve as new memories are added~\cite{xu2025amem}.
AgentRefine instead learns correction behavior from environment-feedback trajectories through refinement tuning~\cite{fu2025agentrefine}.
AgentGym supplies diverse interactive environments and trajectory sets for evaluating and training agents through continued interaction~\cite{xi2025agentgym}.
JitRL retrieves state--action--return experience to estimate action advantages and adjusts policy logits without gradient updates~\cite{li2026jitrl}.
Together, these methods distinguish memory organization, parameter learning, and inference-time policy correction as complementary routes to adaptation.
For a physical swarm, reward-based reuse also requires relating experience to the local state and responsibility of the agent making the current decision.

Connecting language-mediated choices to external tools and physical actuators places additional demands on the execution interface.
AgentDojo evaluates prompt-injection attacks and defenses when agents invoke tools over untrusted data~\cite{debenedetti2024agentdojo}.
BadRobot shows how language-model vulnerabilities can propagate into physical actions in embodied systems~\cite{zhang2025badrobot}.
Thea formulates the surrounding context, action interface, and execution feedback as an embodied-agent harness~\cite{wang2026thea}.
SHAPER further studies the joint evolution of reusable skills and a context--code harness around a frozen language model~\cite{wang2026shaper}.
These studies provide foundations for skill grounding, experience reuse, and execution-interface design.
Their common focus on an individual agent, a shared tool environment, or model-level adaptation leaves the organization of execution authority and experience attribution across concurrently acting physical agents comparatively underexplored.

\subsection{LLM-Based Multi-Agent and Multi-Robot Systems}

LLM-based multi-agent research extends language-mediated reasoning from a single embodied agent to teams that divide tasks, exchange information, and coordinate actions toward a shared objective.
SMART-LLM separates task decomposition, coalition formation, and allocation to generate multi-robot task plans~\cite{kannan2024smartllm}.
CaPo constructs a cooperative meta-plan and revises it through multi-agent discussion as task progress changes~\cite{liu2025capo}.
EMOS incorporates embodiment-derived capability descriptions into hierarchical planning for heterogeneous robot teams~\cite{chen2025emos}.
These systems emphasize task-level organization and capability-aware assignment through shared or hierarchical planning.

Other work places greater emphasis on interaction among individual decision processes.
CoELA combines perception, memory, communication, planning, and execution for cooperation under decentralized control and costly communication~\cite{zhang2024coela}.
RoCo equips robots with language-model agents that negotiate subtask plans and waypoints through dialogue, using motion-planning feedback to revise their proposals~\cite{mandi2024roco}.
PARTNR evaluates embodied collaboration under spatial, temporal, and heterogeneous capability constraints.
However, the evaluated agents still struggle to coordinate their actions, track task progress, and recover from errors~\cite{chang2025partnr}.
These findings highlight a gap between generating cooperative plans and maintaining coordinated execution as a task unfolds.

\begin{figure*}[!t]
  \centering
  \includegraphics[width=\textwidth]{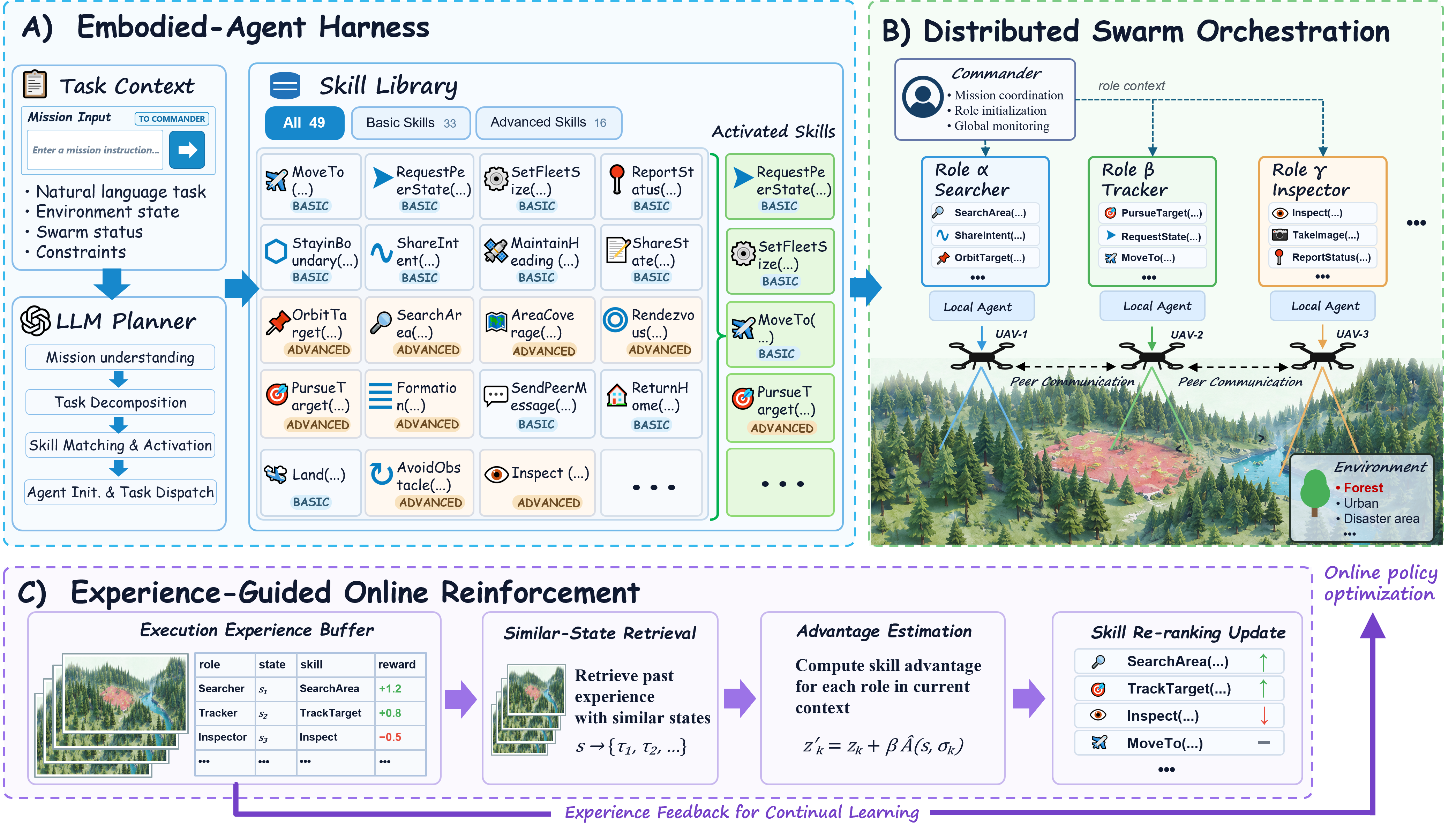}
  \caption{\textbf{Overview of AeroWeaver.} The embodied-agent harness \textbf{(A)} maps task context to a subset of Aerial Skills. Distributed swarm orchestration \textbf{(B)} instantiates role-conditioned local agents, binds each agent to one UAV, and supports directed peer coordination during execution. Experience-guided online reinforcement \textbf{(C)} retrieves role-relevant trajectories, estimates skill advantages from rewards, and updates the active skill ranking.}
  \label{fig:aeroweaver-framework}
\end{figure*}

\ifdefined\ARXIVVERSION
\begin{figure}[!b]
  \centering
  \includegraphics[width=\columnwidth]{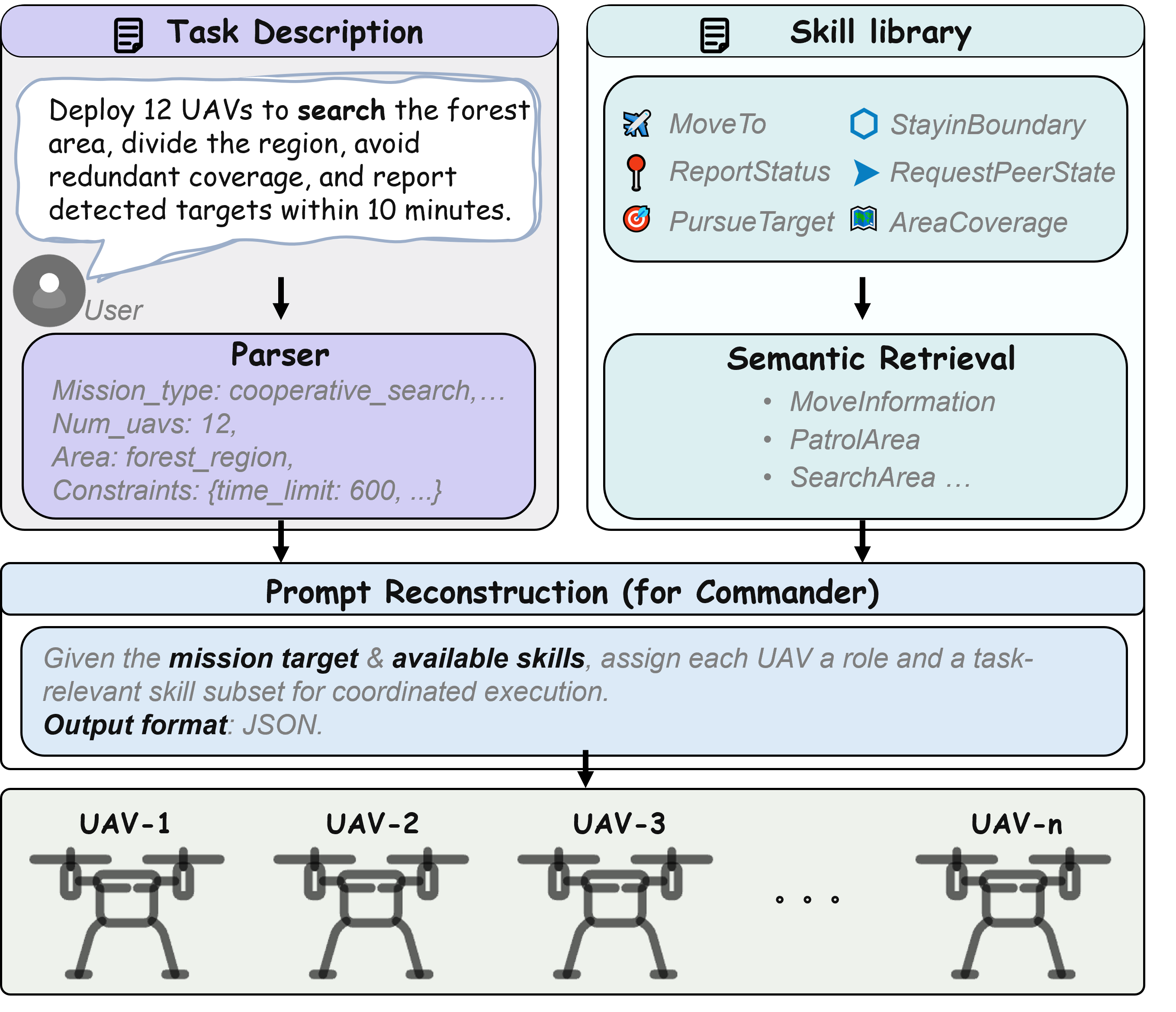}
  \caption{\textbf{Task-conditioned skill activation.} Mission parsing and skill retrieval construct the Commander context for role assignment and activation of agent-specific skill subsets.}
  \label{fig:skill-library}
\end{figure}

\begin{figure*}[!t]
  \centering
  \includegraphics[width=0.9\linewidth]{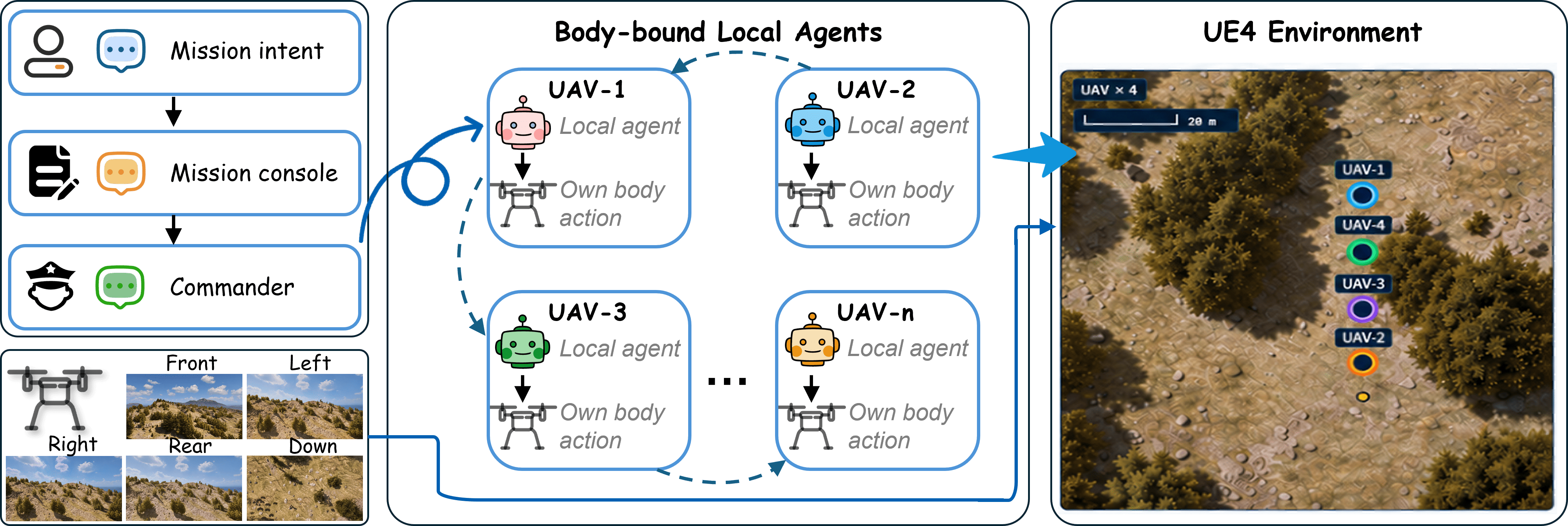}
  \caption{\textbf{Distributed swarm execution and communication.} The Commander supplies mission context to UAV-bound agents that select own-body actions and exchange peer messages; camera views and a four-UAV UE4 scene illustrate the deployment.}
  \label{fig:role-orchestration}
\end{figure*}
\fi

Research on aerial teams has adapted language-model reasoning to UAV-specific capabilities, mission interfaces, and learning processes.
FlockGPT maps natural-language descriptions to UAV formation geometry~\cite{flockgpt}.
TALKER activates reusable action primitives and maintains an extensible knowledge library for multi-UAV missions~\cite{lou2025talker}.
Agents Trainer uses cooperating language-model agents to automate multi-agent reinforcement-learning configuration, reward design, and policy training for drone swarms~\cite{lou2026agents}.
AERIS dynamically rebinds role-specialized language modules across aerial executors at runtime~\cite{lou2026aeris}.
These systems broaden the role of language models in aerial teams from mission interpretation to skill reuse, coordination, interface grounding, and policy development.
Across their reported designs, mission reasoning, vehicle execution, and learning are typically implemented as distinct stages or subsystems.
Experience reuse has also been studied at the level of multi-agent orchestration.
Skill-MAS treats orchestration knowledge as a non-parametric object that can be refined from multiple trajectories~\cite{lin2026skillmas}.

Taken together, prior work provides strong foundations for grounded skills, agent adaptation, and language-mediated team coordination.
Integrating these capabilities in a physical swarm remains a systems question because semantic decisions, observations, communication, and actuators are distributed across different components.

\section{Method}
\label{sec:method}
AeroWeaver is an embodied-agent harness that weaves Aerial Skills into coordinated UAV swarm execution.
As shown in Fig.~\ref{fig:aeroweaver-framework}, AeroWeaver integrates three components:
\begin{itemize}[\labelindent=0pt]
    \item \textbf{(A) Aerial Skill Harness} activates task-relevant Aerial Skills by parsing mission context, retrieving from the skill library, and exposing an active skill subset to each role-conditioned agent (Sec.~\ref{sec:governed-harness}).
    \item \textbf{(B) Distributed Swarm Orchestration} maintains distributed execution through role-conditioned local agents that coordinate via directed peer messages (Sec.~\ref{sec:distributed-orchestration}).
    \item \textbf{(C) Experience-Guided Reinforcement} enables training-free online adaptation by refining agent policy from execution experience (Sec.~\ref{sec:online-reinforcement}).
\end{itemize}

\subsection{Embodied-Agent Harness with Aerial Skills}

\label{sec:governed-harness}
Within the embodied-agent harness, Aerial Skills constitute an executable intermediate layer between high-level semantic planning and low-level motion control.
This layer exposes each downstream UAV agent to a skill subset while retaining typed interfaces, body bindings, and approved executors within the runtime.
Fig.~\ref{fig:skill-library} illustrates how skill activation instantiates this interface.

\ifdefined\ARXIVVERSION
\else
\begin{figure}[h]
  \centering
  \includegraphics[width=\columnwidth]{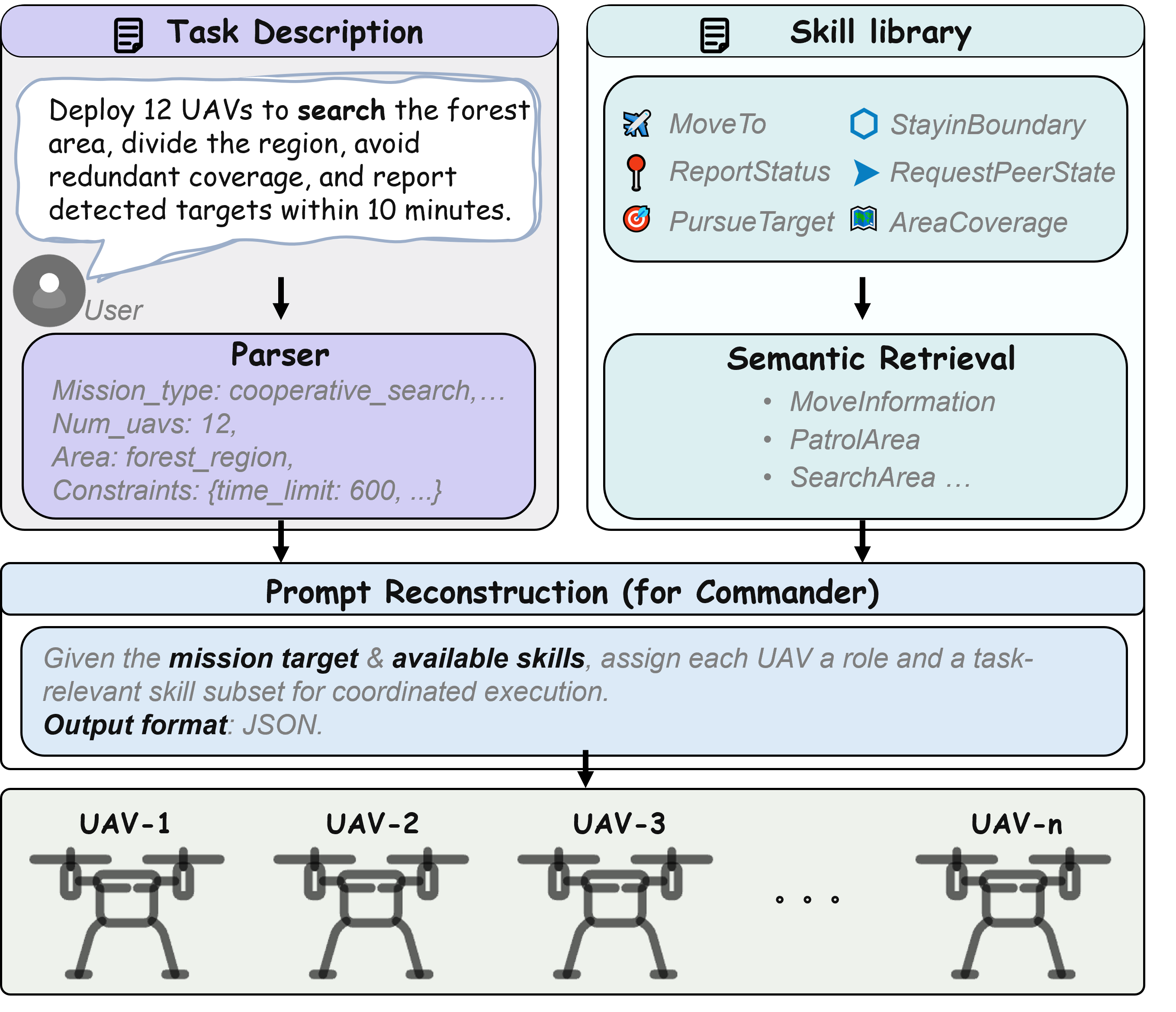}
  \caption{\textbf{Task-conditioned skill activation.} Mission parsing and skill retrieval construct the Commander context for role assignment and activation of agent-specific skill subsets.}
  \label{fig:skill-library}
\end{figure}

\begin{figure*}[h]
  \centering
  \includegraphics[width=0.9\linewidth]{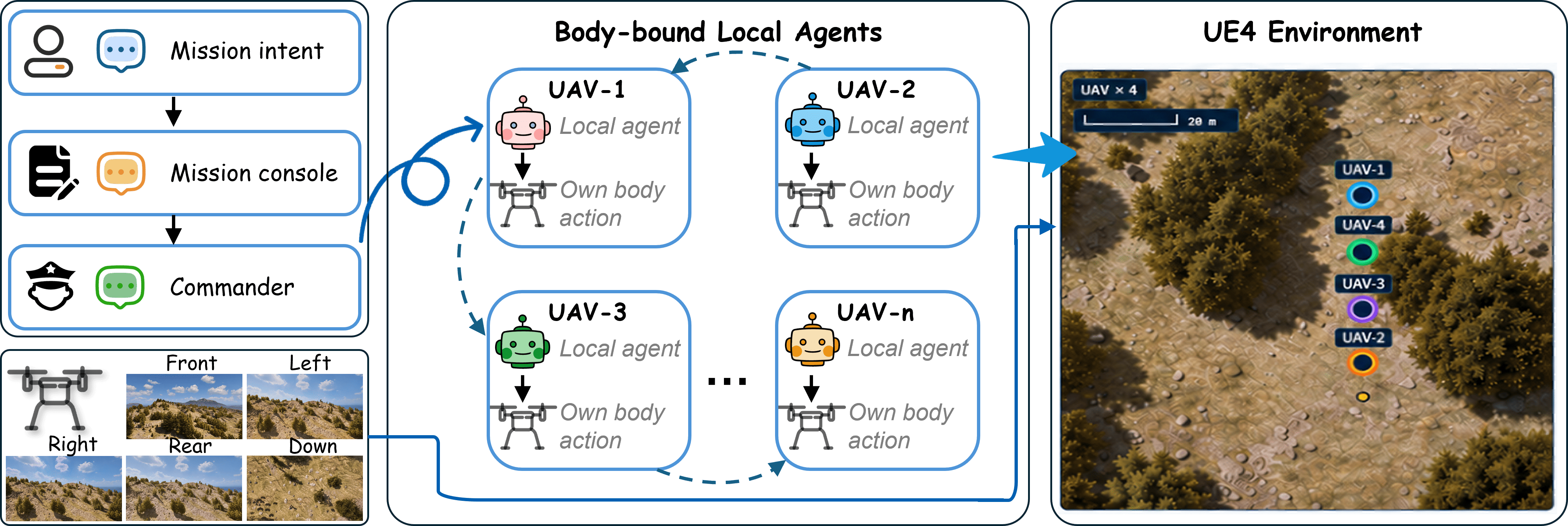}
  \caption{\textbf{Distributed swarm execution and communication.} The Commander supplies mission context to UAV-bound agents that select own-body actions and exchange peer messages; camera views and a four-UAV UE4 scene illustrate the deployment.}
  \label{fig:role-orchestration}
\end{figure*}
\fi

\noindent\textbf{Skill substrate.}
We define a Skill as a reusable capability unit that couples an executable object with a \texttt{skill.md} document describing its semantic purpose, invocation interface, and operating conditions.
Skills are derived from built-in flight and payload functions, platform adapters, perception modules, or compositions of existing skills.
Each skill $\sigma_i$ follows the common representation
\begin{equation}
\sigma_i=\langle n_i,d_i,I_i,O_i,P_i,E_i,M_i\rangle,
\label{eq:skill}
\end{equation}
where $n_i$ and $d_i$ identify and describe the skill, $I_i$ and $O_i$ specify typed inputs and outputs, $P_i$ states its operating conditions, $E_i$ denotes the associated executor, and $M_i$ stores provenance, dependencies, and effects.

\noindent\textbf{Mission parsing and retrieval.}
Given an operator instruction $q$ and swarm state $s_t$, the mission parser extracts a structured task record $\chi^t$ containing the task type, participating UAVs, operating region, and mission constraints.
This record supplies the semantic and operational context used to retrieve skills from the registered library.
For a given task, the harness retrieves a task-level candidate catalog $\mathcal{A}^t$ from the available skills.
The retriever then ranks this catalog by comparing the task and role query with the names, descriptions, tags, and aliases stored in each \texttt{skill.md} document.

\noindent\textbf{Prompt reconstruction and role assignment.}
The harness combines $q$, $\chi^t$, the current swarm state, and the documentation of the task-level candidate catalog to construct the Commander context shown in Fig.~\ref{fig:skill-library}.
The Commander assigns a semantic role $\rho_k$ and local goal $g_k$ to each participant, after which the harness forms the role-conditioned query $q_k=\mathrm{Pack}(q,\rho_k,g_k)$ and activates
\begin{equation}
\mathcal{L}_k=\operatorname{TopK}_{\sigma_i\in\mathcal{A}^t}
\{r_k(q_k,\sigma_i)\},
\label{eq:active-skill-set}
\end{equation}
where $r_k$ is the retrieval score for UAV $k$.
The resulting $\mathcal{L}_k$ determines the skill names and documentation rendered into agent $k$'s local context, and different agents may receive different subsets of the same task-level catalog according to their assigned roles and local goals.


\subsection{Distributed Swarm Orchestration}
\label{sec:distributed-orchestration}
Distributed Swarm Orchestration maintains a shared mission through concurrent decision processes bound to individual UAVs. As shown in Fig.~\ref{fig:role-orchestration}, task context establishes their local objectives, directed peer messages support coordination, and local reports provide mission-progress feedback. The figure depicts $n$ body-bound agents alongside a four-UAV UE4 scene, with front, left, right, rear, and downward camera views shown in the lower-left panel.

\noindent\textbf{Body-bound local decisions.}
Following role assignment and skill activation in Sec.~\ref{sec:governed-harness}, agent $k$ is bound to UAV $u_k$ and receives the active skill subset $\mathcal{L}_k$.
Its context $c_k^t$ contains the assigned role, local goal, mission constraints, skill documentation, and an isolated interaction history.
At round $t$, the agent selects a parameterized skill invocation
\begin{equation}
a_k^t=\pi_{\rm sem}\!\left(c_k^t,o_k^t,
\{m_{\ell\rightarrow k}^{t-1}\}_{\ell\in\mathcal{N}_k^t},\mathcal{L}_k\right),
\label{eq:distributed-decision}
\end{equation}
where $o_k^t$ is the body-local observation and $\mathcal{N}_k^t$ identifies the currently reachable peers.

\noindent\textbf{Directed peer communication.}
Communication supplements body-local sensing with information from reachable peers.
The neighborhood $\mathcal{N}_k^t$ contains other UAVs within the configured communication radius $r_{\rm comm}$ and changes as the vehicles move.
A message $m_{k\rightarrow\ell}^t$ names its sender and recipient and carries role status, local observations, intended motion, or a request for peer state.
Received messages enter the recipient's next-round decision context without transferring execution authority between agents.
\begin{figure*}[h]
  \centering
  \includegraphics[width=0.95\linewidth]{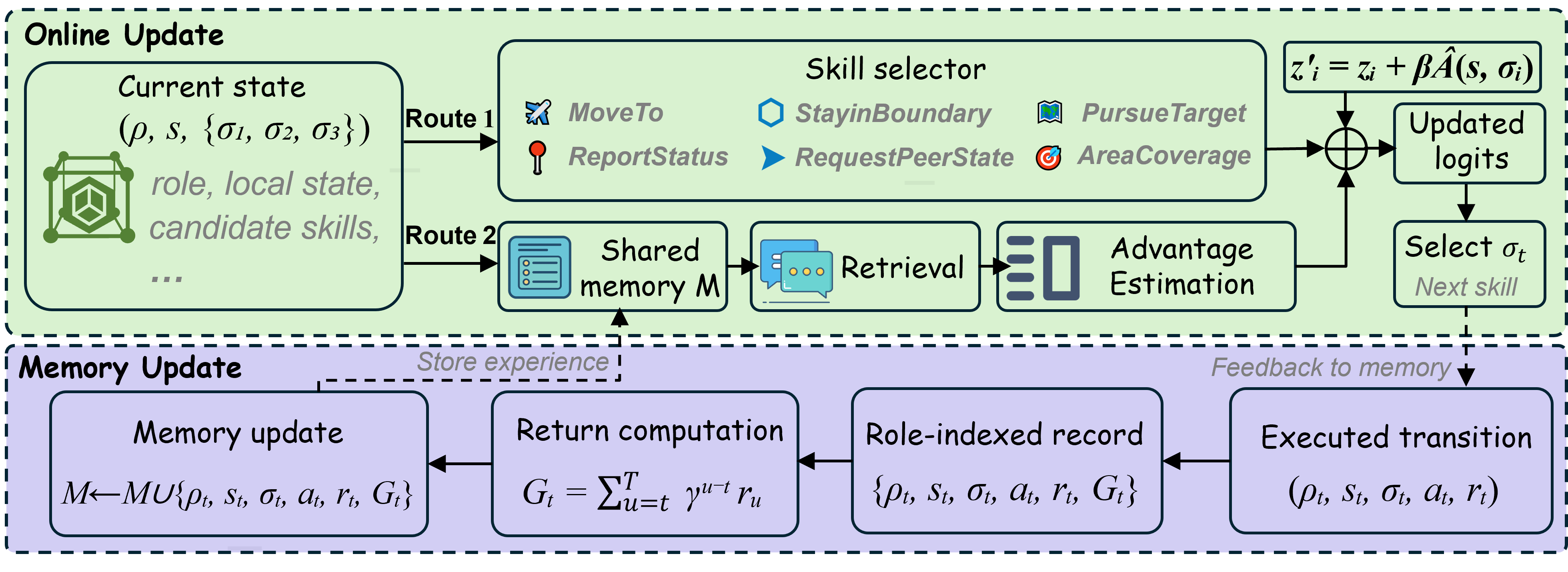}
  \vspace{-8pt}
  \caption{\textbf{Experience-guided online reinforcement.} Retrieved reward evidence adjusts skill selection (upper stage), while executed transitions update role-indexed memory (lower stage).}
  \label{fig:memory-reinforcement}
\end{figure*}

\noindent\textbf{Execution and progress feedback.}
Each world round refreshes local observations and communication neighborhoods, runs the agents' decisions concurrently, and applies their selected invocations to the bound UAVs.
In the pursuit runtime, each UAV retains its previous velocity until its own agent selects a new direction.

The Commander supplies task context and monitors mission progress, with no flight authority.
Local role reports follow a single supervisory path through the Commander and Mission Console to the user.
The local execution records supply role-indexed experience for the online reinforcement mechanism in Sec.~\ref{sec:online-reinforcement}.

\subsection{Experience-Guided Online Reinforcement}
\label{sec:online-reinforcement}

Execution experience provides reward evidence for adapting the skill preferences of local agents.
As shown in Fig.~\ref{fig:memory-reinforcement}, AeroWeaver combines an online update that adjusts current selector scores (upper stage) with a memory update that associates executed skills with their observed returns (lower stage). Dashed connectors denote execution feedback and subsequent experience reuse.
The resulting reinforcement operates over the active skill set and accumulates across decision rounds through shared swarm memory.

\noindent\textbf{Online update.}
For a local decision, let $\rho$ denote the agent's role, $s$ its current local state, and $\mathcal{L}$ the active skill set supplied by the harness.
The state summarizes local observations, received peer messages, execution progress, and relevant mission context.
Each callable skill--parameter pair receives a single-token label; its provider-reported first-token log probability defines $z_i$. Parameterizations have separate base scores but share a skill advantage. Up to 20 alternative scores are requested. A corrected choice is used only when omitted candidates cannot exceed it under the returned probability bound; otherwise the highest observed base score is retained.
Memory retrieval filters reusable records by exact task and role, then retains up to 32 by Jaccard overlap of tokenized state descriptions, breaking ties by recency. Baseline and per-skill returns are unweighted means; different bodies may contribute under the same role.

Let $\mathcal{N}_{\rho}(s)$ denote the retrieved neighborhood, $\bar G_{\rho}(s)$ its mean return, and $\bar G_{i,\rho}(s)$ the mean return of records associated with skill $\sigma_i$.
For the current role, the estimated skill advantage is
\begin{equation}
\widehat A(s,\sigma_i)=\bar G_{i,\rho}(s)-\bar G_{\rho}(s).
\label{eq:skill-advantage}
\end{equation}
A positive estimate raises the candidate's score, while a negative estimate lowers it.
When no matching experience is available for a candidate, its correction is set to zero.

The additive correction adjusts the scores of active skills:
\begin{equation}
z'_i=z_i+\beta\widehat A(s,\sigma_i),\qquad \beta\geq 0,
\label{eq:online-adaptation}
\end{equation}
where $\beta$ controls the influence of retrieved reward evidence on the selector's initial preference.
The agent selects the highest-scoring skill $\sigma_t$ from the updated ranking and dispatches its invocation through the existing body-bound execution interface.
The candidate set remains $\mathcal{L}$ throughout this update, and $\beta=0$ recovers the base selector.

\noindent\textbf{Memory update.}
Executing the selected skill produces a transition summarized by $(\rho_t,s_t,\sigma_t,a_t,r_t)$ in the lower part of Fig.~\ref{fig:memory-reinforcement}.
Here $\sigma_t$ identifies the skill, $a_t$ is its concrete invocation with execution parameters, and $r_t$ is the task-environment reward for the executing participant.
The reward is stored unchanged, with benchmark-specific task and reward definitions given in Sec.~\ref{sec:experiments}.
Rewards are accumulated along that participant's trajectory to associate each skill choice with its subsequent outcomes,
\begin{equation}
G_t=\sum_{u=t}^{T}\gamma^{u-t}r_u,
\label{eq:experience-return}
\end{equation}
where $\gamma\in[0,1]$ is the discount factor and $T$ is the last observed transition in the stored trajectory segment.
As additional rewards arrive, the returns of earlier decisions are extended and are finalized when the trajectory ends.
Each online selection therefore uses only reward evidence already observed before that decision.

Memory stores each outcome as a role-indexed record $(\rho_t,s_t,\sigma_t,a_t,r_t,G_t)$.
Vehicle and trajectory identifiers retain the provenance of each record, while role and state determine its relevance to later queries.
This cycle supports training-free adaptation through accumulated execution experience while keeping model parameters, skill definitions, and flight executors fixed.

\section{Experiments}
\label{sec:experiments}

The evaluation covers nine closed-loop tasks, two LLM baselines, and three component controls.
The comparison covers 54 task--condition pairs with 10 episode-return observations per pair.

\subsection{Experimental Setup}
\label{sec:experimental-setup}

\begin{figure}[h]
  \centering
  \captionsetup[subfloat]{font=mpepanel,labelformat=parens,labelsep=space,justification=centering,singlelinecheck=true,margin=0pt,captionskip=1.5pt,farskip=0pt,nearskip=0pt,topadjust=0pt}
  \subfloat[Coverage\label{fig:mpe-coverage}]{%
    \includegraphics[width=0.33\columnwidth]{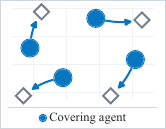}}\hfill
  \subfloat[Pursuit-evasion\label{fig:mpe-pursuit}]{%
    \includegraphics[width=0.33\columnwidth]{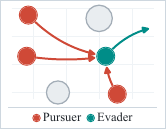}}\hfill
  \subfloat[Guided navigation\label{fig:mpe-navigation}]{%
    \includegraphics[width=0.33\columnwidth]{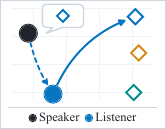}}
  \par\vspace{2pt}
  \subfloat[Private communication\label{fig:mpe-private-communication}]{%
    \includegraphics[width=0.33\columnwidth]{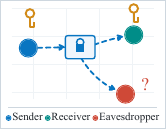}}\hfill
  \subfloat[Circular formation\label{fig:mpe-circular-formation}]{%
    \includegraphics[width=0.33\columnwidth]{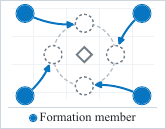}}\hfill
  \subfloat[Line formation\label{fig:mpe-line-formation}]{%
    \includegraphics[width=0.33\columnwidth]{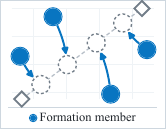}}
  \par\vspace{2pt}
  \subfloat[World communication\label{fig:mpe-world-communication}]{%
    \includegraphics[width=0.33\columnwidth]{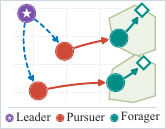}}\hfill
  \subfloat[Collection-delivery\label{fig:mpe-collection-delivery}]{%
    \includegraphics[width=0.33\columnwidth]{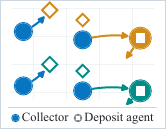}}\hfill
  \subfloat[Goal concealment\label{fig:mpe-goal-concealment}]{%
    \includegraphics[width=0.33\columnwidth]{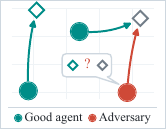}}
  \par\vspace{1pt}
  \includegraphics[width=\columnwidth]{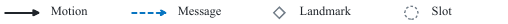}
  \caption{\textbf{MPE-inspired task forms and roles.} Nine tasks span spatial coordination, communication, and cooperative--competitive interaction; solid and dashed arrows denote motion and information exchange.}
  \label{fig:mpe-task-forms}
\end{figure}

\noindent\textbf{Environment and implementation.}
We evaluate nine MPE-inspired tasks~\cite{lowe2017multiagent}, illustrated in Fig.~\ref{fig:mpe-task-forms}. Coverage and circular/line formation assess spatial coordination, while guided navigation, private communication, and world communication involve role-dependent information exchange. Pursuit-evasion tests coordinated pursuit, collection-delivery requires cooperative collection and delivery, and goal concealment combines target reaching with hiding the goal from an adversary. Task rewards are computed using MPE2 scenario functions evaluated on our simulation runtime states.

All methods share an AirSim~\cite{airsim} simulation runtime, task rewards, and fixed opponent policies. The area is $140\times140$\,m at fixed altitude; default neighbor-sensing and communication ranges are 45\,m and 55\,m, with a 100\,m sensing range for the world-communication leader. Local observations expose own state, visible entities, received messages, and callable skill--parameter pairs; private information remains role-restricted. Agents select from a frozen round state, then the world advances by 0.5\,s. Invalid invocations produce an own-body hold. 
The evaluation uses 22 task skills; the deployment catalog in Fig.~\ref{fig:aeroweaver-framework} additionally contains platform-specific capabilities.
We use DeepSeek V4-flash~\cite{deepseekai2026deepseekv4} as the LLM backbone for the main comparisons.
Experience correction uses $\gamma=0.95$ and $\beta=0.8$, without return or advantage rescaling.

Episodes run for at most 24 rounds, with task-specific early termination. Coverage, navigation, and formation episodes end after three rounds within 3\,m of all targets; pursuit requires two pursuers within 6\,m of the evader. Private communication ends after three rounds, collection after all deliveries, and world communication and concealment at the horizon.

\noindent\textbf{Baselines.}
We compare AeroWeaver with two LLM-based baselines.
\textbf{Centralized} is an in-house planner that aggregates permitted team observations and directly selects skill calls for all controlled participants at each step.
\textbf{HMAS-2}, adapted to our tasks from~\cite{chen2024scalable}, first generates a central plan; local agents review their assigned actions using local observations, and the central planner revises the plan once based on their feedback before execution.
Both baselines receive all 22 skills without reward memory, sharing AeroWeaver's action constraints and skill executors.
Component ablations separately disable skill activation, coordination reports, or reward correction.

\noindent\textbf{Evaluation metrics.}
Task performance is measured by the undiscounted episode return, $R=|C|^{-1}\sum_{i\in C}\sum_{t=1}^{T}r_{i,t}$, where $C$ is the set of controlled agents and $T$ is the episode length.
We report the mean and sample standard deviation of raw returns over 10 episodes per task--condition pair.
Inference cost is measured by provider-reported token consumption per episode, including decision calls and, where applicable, one full skill-activation setup.

\begin{table}[h]
\centering
\caption{End-to-end episode rewards (mean $\pm$ SD; $n=10$).}
\label{tab:end-to-end-pilot}
\footnotesize
\setlength{\tabcolsep}{1.4pt}
\renewcommand{\arraystretch}{1.12}
\begin{tabular*}{\columnwidth}{@{\extracolsep{\fill}}lrrr@{}}
\toprule
Task & Centralized & HMAS-2 & AeroWeaver \\
\midrule
Coverage & \textminus{}14.81\,\textpm\,1.02 & \textminus{}14.81\,\textpm\,1.00 & \textbf{\textminus{}14.20\,\textpm\,0.95} \\
Pursuit-evasion & 90.00\,\textpm\,5.00 & 70.00\,\textpm\,4.60 & \textbf{130.00\,\textpm\,4.80} \\
Guided navigation & \textminus{}8.53\,\textpm\,0.36 & \textminus{}8.53\,\textpm\,0.35 & \textbf{\textminus{}7.10\,\textpm\,0.32} \\
Private communication & \textminus{}2.00\,\textpm\,0.11 & \textminus{}2.00\,\textpm\,0.10 & \textbf{\textminus{}1.85\,\textpm\,0.09} \\
Circular formation & \textminus{}8.45\,\textpm\,0.12 & \textminus{}8.45\,\textpm\,0.11 & \textbf{\textminus{}8.20\,\textpm\,0.10} \\
Line formation & \textminus{}13.99\,\textpm\,0.27 & \textminus{}7.62\,\textpm\,0.23 & \textbf{\textminus{}7.30\,\textpm\,0.22} \\
World communication & 129.00\,\textpm\,5.30 & 115.00\,\textpm\,0.90 & \textbf{135.00\,\textpm\,5.20} \\
Collection-delivery & 33.91\,\textpm\,0.85 & 38.70\,\textpm\,0.78 & \textbf{49.50\,\textpm\,0.68} \\
Goal concealment & \textminus{}1.06\,\textpm\,0.14 & \textminus{}0.41\,\textpm\,0.10 & \textbf{\textminus{}0.25\,\textpm\,0.11} \\
\bottomrule
\end{tabular*}
\end{table}

\subsection{Main Results}
\label{sec:main-results}

\noindent\textbf{Task performance.}
AeroWeaver achieves the highest mean episode reward on all nine tasks in Table~\ref{tab:end-to-end-pilot}, outperforming Centralized and HMAS-2 across spatial coordination, communication, and cooperative--competitive interaction.
Centralized aggregates permitted team observations and the full skill catalog into a long context, then jointly assigns skills and body-specific parameters in one response, increasing the information and assignment burden of each decision.
HMAS-2 adds local review, but its advantage over Centralized varies across tasks; neither baseline uses reward memory.
AeroWeaver instead organizes selection around task-relevant skills and each agent's executable options, reducing the number of capabilities and role-specific assignments considered in each decision.
Peer reports inform complementary actions, while retrieved reward evidence updates skill preferences as execution proceeds.
The consistently higher returns are compatible with this combination of focused selection, distributed coordination, and experience feedback, which connects mission objectives to local action choices throughout an episode.

\begin{figure}[h]
\centering
\includegraphics[width=\columnwidth]{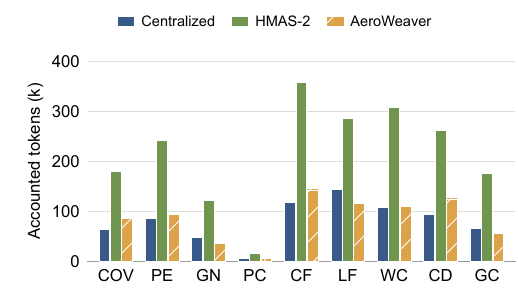}
\caption{\textbf{Token consumption.} Per-episode totals for Centralized, HMAS-2, and AeroWeaver include activation setup and decision calls; lower is better.}
\label{fig:main-token-comparison}
\end{figure}

\begin{figure*}[!h]
\centering
\captionsetup[subfloat]{font=footnotesize,labelformat=parens,labelsep=space,justification=centering,singlelinecheck=true,captionskip=2pt,farskip=0pt,nearskip=0pt}
\subfloat[Experience-guided adaptation\label{fig:additional-adaptation}]{%
  \includegraphics[width=0.33\textwidth]{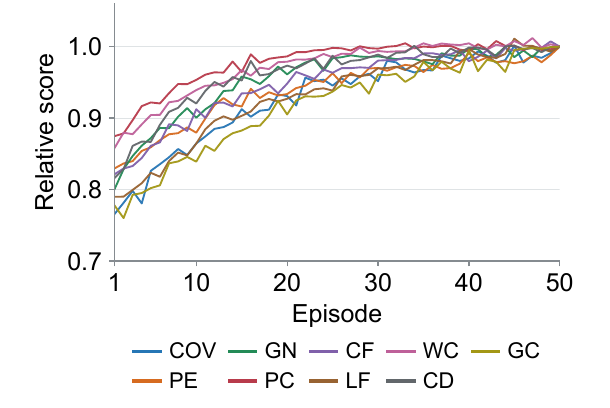}}\hfill
\subfloat[Backbone sensitivity\label{fig:additional-backbone}]{%
  \includegraphics[width=0.33\textwidth]{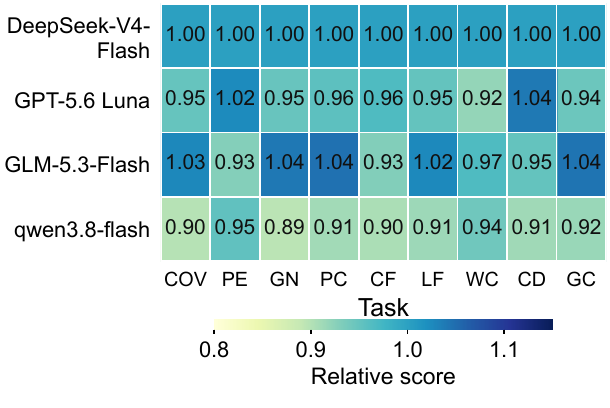}}\hfill
\subfloat[Prompt perturbation sensitivity\label{fig:additional-prompt}]{%
  \includegraphics[width=0.33\textwidth]{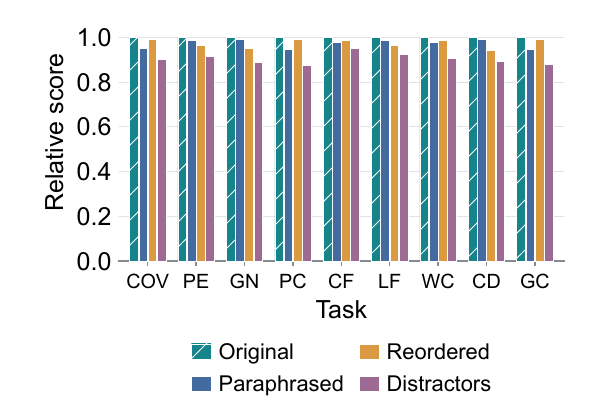}}
\caption{\textbf{Additional experiments.} Panels illustrate adaptation trajectories and task-wise comparisons of LLM backbones and prompt perturbations.}
\label{fig:additional-preview}
\end{figure*}

\noindent\textbf{Token efficiency.}
As shown in Fig.~\ref{fig:main-token-comparison}, AeroWeaver consumes 52.0--70.0\% fewer tokens than HMAS-2 across the nine tasks and fewer tokens than Centralized on four tasks.
Task-conditioned activation limits the skill documentation processed at each decision, while local action selection eliminates the central planning and revision calls required by HMAS-2.
Relative to Centralized, the shorter local contexts are offset by separate calls for individual agents, making the overall token advantage task-dependent.
The reported totals include activation setup and all decision calls from one episode per condition, with task-dependent termination and episode lengths.

\subsection{Ablation Studies}
\label{sec:ablation-studies}

Table~\ref{tab:component-ablations} evaluates the removal of skill activation, coordination reports, and reward correction. The controls respectively expose the full catalog, suppress coordination reports while preserving task-required messages, and set $\beta=0$. For task $t$ and variant $m$, the reported relative mean score is
\begin{equation}
Q_{t,m}=1+\frac{\bar R_{t,m}-\bar R_{t,\mathrm{full}}}{K_t},\qquad K_t>0,
\label{eq:ablation-relative-score}
\end{equation}
where $\bar R$ is the mean episode return and $K_t$ is the fixed task-specific scale of the existing affine score transformation, shared by all variants. AeroWeaver therefore scores 1, and lower scores indicate lower returns. Table~\ref{tab:component-ablations} lists the scales, so raw means are recovered as $\bar R_{t,m}=\bar R_{t,\mathrm{full}}+K_t(Q_{t,m}-1)$.

\begin{table}[h]
\centering
\caption{Component ablations across nine tasks.}
\label{tab:component-ablations}
\footnotesize
\setlength{\tabcolsep}{1.5pt}
\renewcommand{\arraystretch}{1.15}
\begin{tabular*}{\columnwidth}{@{\extracolsep{\fill}}lccc@{}}
\toprule
Task & \shortstack{w/o skill\\activation} & \shortstack{w/o coordination\\reports} & \shortstack{w/o reward\\correction} \\
\midrule
Coverage & 0.650 & 0.708 & 0.764 \\
Pursuit-evasion & 0.670 & 0.781 & 0.823 \\
Guided navigation & 0.734 & 0.632 & 0.802 \\
Private communication & 0.812 & 0.617 & 0.874 \\
Circular formation & 0.762 & 0.763 & 0.825 \\
Line formation & 0.722 & 0.821 & 0.798 \\
World communication & 0.790 & 0.633 & 0.854 \\
Collection-delivery & 0.632 & 0.628 & 0.814 \\
Goal concealment & 0.777 & 0.605 & 0.780 \\
\bottomrule
\end{tabular*}
\par\vspace{4pt}
\begin{minipage}{\columnwidth}
\scriptsize
Relative mean scores ($n=10$ per condition), normalized to AeroWeaver within each task; the full model has a reference score of 1. $K_t$ in row order: 17.077, 17.606, 15.029, 15.969, 16.801, 14.280, 35.217, 10.881, and 13.505.
\end{minipage}
\end{table}

Removing coordination reports produces the largest decrease in five tasks: guided navigation, private communication, world communication, collection-delivery, and goal concealment. The remaining four tasks show the largest decrease when skill activation is removed. Goal concealment is particularly sensitive to coordination reports, with a score of 0.605 compared with 0.777 without activation and 0.780 without reward correction. Disabling reward correction yields scores of 0.764--0.874 and is the least detrimental ablation in eight tasks; line formation is the exception, where removing coordination reports retains a higher score. These comparisons indicate task-dependent contributions from activation and coordination, with reward correction providing an additional improvement across the suite.

\subsection{Additional Experiments}
\label{sec:additional-experiments}

The additional studies address experience reuse, dependence on the LLM backbone, and sensitivity to mission wording. Fig.~\ref{fig:additional-preview}(a)--(c) illustrates the measured results.

\noindent\textbf{(a) Experience-guided adaptation.}
Fig.~\ref{fig:additional-preview}(a) shows rapid early improvement followed by gradual stabilization across the nine tasks. This trend is consistent with experience-guided online adaptation, as accumulated reward information refines subsequent skill selection without changing model weights or skill executors.

\noindent\textbf{(b) Backbone sensitivity.}
In Fig.~\ref{fig:additional-preview}(b), GPT-5.6 Luna and GLM-5.3-Flash remain close to the DeepSeek-V4-Flash reference of 1 across the nine tasks, while qwen3.8-flash exhibits a moderate performance decrease. Overall, AeroWeaver maintains relatively stable performance across different LLM backbones, demonstrating robustness to backbone choice despite task-dependent variations.

\noindent\textbf{(c) Prompt perturbation sensitivity.}
The prompt comparison varies only the mission text, keeping the activated skill set fixed. Paraphrasing preserves task meaning, reordering reverses sentence order, and distractor insertion appends unrelated report metadata without changing task requirements. With Original normalized to 1 within each task, Fig.~\ref{fig:additional-preview}(c) depicts smaller score reductions under paraphrasing and reordering than under distractor insertion, illustrating a stronger response to irrelevant context than to changes in wording or order.

\section{Conclusion}
\label{sec:conclusion}

We introduced AeroWeaver, an embodied-agent harness for UAV swarms that integrates task-conditioned skill activation, distributed body-local coordination, and role-indexed experience reuse within a unified execution framework. Reward-derived evidence is incorporated into subsequent skill selection without modifying the underlying model parameters or low-level flight controllers. Across the nine evaluated tasks, AeroWeaver achieved higher mean episodic returns than Centralized and HMAS-2. It also required fewer tokens than HMAS-2 in the recorded episodes, while its efficiency relative to Centralized varied across tasks. The ablation results further show that skill activation, coordination reports, and reward-based correction contribute to overall performance.

The current evaluation is conducted with a predefined skill catalog and simulated execution conditions. Future work will extend AeroWeaver toward longer-term experience reuse, broader task transfer, and validation on physical UAV platforms.
\bibliographystyle{IEEEtran}
\bibliography{references}

@IEEEtranBSTCTL{IEEEexample:BSTcontrol,
  CTLuse_forced_etal = {yes},
  CTLmax_names_forced_etal = {6},
  CTLnames_show_etal = {1},
  CTLuse_url = {no}
}

@inproceedings{chen2024scalable,
  title = {Scalable Multi-Robot Collaboration with Large Language Models: Centralized or Decentralized Systems?},
  author = {Chen, Yongchao and Arkin, Jacob and Zhang, Yang and Roy, Nicholas and Fan, Chuchu},
  booktitle = {Proc. IEEE Int. Conf. Robot. Autom. (ICRA)},
  year = {2024},
  url = {https://arxiv.org/abs/2309.15943},
  pages = {4311--4317},
}

@inproceedings{flockgpt,
  title = {{FlockGPT}: Guiding {UAV} Flocking with Linguistic Orchestration},
  author = {Lykov, Artem and Karaf, Sausar and Martynov, Mikhail and Serpiva, Valerii and Fedoseev, Aleksey and Konenkov, Mikhail and Tsetserukou, Dzmitry},
  booktitle = {Proc. IEEE Int. Symp. Mixed Augmented Reality Adjunct (ISMAR-Adjunct)},
  pages = {485--488},
  year = {2024},
}

@inproceedings{saycan,
  title = {Do As {I} Can, Not As {I} Say: Grounding Language in Robotic Affordances},
  author = {Ichter, Brian and Brohan, Anthony and Chebotar, Yevgen and Finn, Chelsea and Hausman, Karol and Herzog, Alexander and Ho, Daniel and Ibarz, Julian and Irpan, Alex and Jang, Eric and Julian, Ryan and Kalashnikov, Dmitry and Levine, Sergey and Lu, Yao and Parada, Carolina and Rao, Kanishka and Sermanet, Pierre and Toshev, Alexander T and Vanhoucke, Vincent and Xia, Fei and Xiao, Ted and Xu, Peng and Yan, Mengyuan and Brown, Noah and Ahn, Michael and Cortes, Omar and Sievers, Nicolas and Tan, Clayton and Xu, Sichun and Reyes, Diego and Rettinghouse, Jarek and Quiambao, Jornell and Pastor, Peter and Luu, Linda and Lee, Kuang-Huei and Kuang, Yuheng and Jesmonth, Sally and Joshi, Nikhil J. and Jeffrey, Kyle and Ruano, Rosario Jauregui and Hsu, Jasmine and Gopalakrishnan, Keerthana and David, Byron and Zeng, Andy and Fu, Chuyuan Kelly},
  booktitle = {Proc. Conf. Robot Learn. (CoRL)},
  pages = {287--318},
  year = {2023},
  volume = {205},
  series = {PMLR},
  url = {https://proceedings.mlr.press/v205/ichter23a.html},
}

@article{Voyager,
  author = {Wang, Guanzhi and Xie, Yuqi and Jiang, Yunfan and Mandlekar, Ajay and Xiao, Chaowei and Zhu, Yuke and Fan, Linxi and Anandkumar, Anima},
  title = {{Voyager}: An Open-Ended Embodied Agent with Large Language Models},
  journal = {Trans. Mach. Learn. Res.},
  year = {2024},
  url = {https://mlanthology.org/tmlr/2024/wang2024tmlr-voyager/},
}

@inproceedings{lowe2017multiagent,
  author = {Ryan Lowe and Yi Wu and Aviv Tamar and Jean Harb and Pieter Abbeel and Igor Mordatch},
  title = {Multi-Agent Actor-Critic for Mixed Cooperative-Competitive Environments},
  booktitle = {Adv. Neural Inf. Process. Syst.},
  volume = {30},
  year = {2017},
  url = {https://proceedings.neurips.cc/paper_files/paper/2017/hash/68a9750337a418a86fe06c1991a1d64c-Abstract.html},
}

@misc{wang2026thea,
  author = {Wang, Qi and Wang, Tianyi and Li, Chengyang and Ban, Shikun and Chen, Yurun and Ge, Yizhong and Qin, Jason and Li, Chengtai and Zhu, Wentao},
  title = {Towards the Harness of Embodied Agents},
  year = {2026},
  note = {arXiv:2608.11246},
  url = {https://arxiv.org/abs/2608.11246},
}

@inproceedings{chen2025emos,
  author = {Chen, Junting and Yu, Checheng and Zhou, Xunzhe and Xu, Tianqi and Mu, Yao and Hu, Mengkang and Shao, Wenqi and Wang, Yikai and Li, Guohao and Shao, Lin},
  title = {{EMOS}: Embodiment-Aware Heterogeneous Multi-Robot Operating System with {LLM} Agents},
  booktitle = {Proc. ICLR},
  year = {2025},
  url = {https://proceedings.iclr.cc/paper_files/paper/2025/hash/8ec46cecdae5407662899c5f6698cd8b-Abstract-Conference.html},
}

@inproceedings{wang2025workflow,
  author = {Wang, Zora Zhiruo and Mao, Jiayuan and Fried, Daniel and Neubig, Graham},
  title = {Agent Workflow Memory},
  booktitle = {Proc. Int. Conf. Mach. Learn. (ICML)},
  series = {PMLR},
  volume = {267},
  pages = {63897--63911},
  year = {2025},
  url = {https://proceedings.mlr.press/v267/wang25bx.html},
}

@inproceedings{fu2025agentrefine,
  author = {Fu, Dayuan and He, Keqing and Wang, Yejie and Hong, Wentao and GongQue, Zhuoma and Zeng, Weihao and Wang, Wei and Wang, Jingang and Cai, Xunliang and Xu, Weiran},
  title = {{AgentRefine}: Enhancing Agent Generalization through Refinement Tuning},
  booktitle = {Proc. ICLR},
  year = {2025},
  url = {https://proceedings.iclr.cc/paper_files/paper/2025/hash/a3cc50126338b175e56bb3cad134db0b-Abstract-Conference.html},
}

@InProceedings{airsim,
author="Shah, Shital
and Dey, Debadeepta
and Lovett, Chris
and Kapoor, Ashish",
editor="Hutter, Marco
and Siegwart, Roland",
title="AirSim: High-Fidelity Visual and Physical Simulation for Autonomous Vehicles",
booktitle="Field and Service Robotics",
year="2018",
publisher="Springer International Publishing",
address="Cham",
pages="621--635",
isbn="978-3-319-67361-5"
}

@inproceedings{xi2025agentgym,
  author = {Xi, Zhiheng and Ding, Yiwen and Chen, Wenxiang and Hong, Boyang and Guo, Honglin and Wang, Junzhe and Guo, Xin and Yang, Dingwen and Liao, Chenyang and He, Wei and Gao, Songyang and Chen, Lu and Zheng, Rui and Zou, Yicheng and Gui, Tao and Zhang, Qi and Qiu, Xipeng and Huang, Xuanjing and Wu, Zuxuan and Jiang, Yu-Gang},
  title = {{AgentGym}: Evaluating and Training Large Language Model-Based Agents across Diverse Environments},
  booktitle = {Proc. Annu. Meeting Assoc. Comput. Linguistics (ACL)},
  pages = {27914--27961},
  year = {2025},
  url = {https://aclanthology.org/2025.acl-long.1355/},
  volume = {1},
}

@inproceedings{debenedetti2024agentdojo,
  author = {Debenedetti, Edoardo and Zhang, Jie and Balunovic, Mislav and Beurer-Kellner, Luca and Fischer, Marc and Tram{\`e}r, Florian},
  title = {{AgentDojo}: A Dynamic Environment to Evaluate Prompt Injection Attacks and Defenses for {LLM} Agents},
  booktitle = {Adv. Neural Inf. Process. Syst.},
  volume = {37},
  year = {2024},
  url = {https://proceedings.neurips.cc/paper_files/paper/2024/hash/97091a5177d8dc64b1da8bf3e1f6fb54-Abstract-Datasets_and_Benchmarks_Track.html},
  pages = {82895--82920},
}

@inproceedings{zhang2025badrobot,
  author = {Zhang, Hangtao and Zhu, Chenyu and Wang, Xianlong and Zhou, Ziqi and Yin, Changgan and Li, Minghui and Xue, Lulu and Wang, Yichen and Hu, Shengshan and Liu, Aishan and Guo, Peijin and Zhang, Leo},
  title = {{BadRobot}: Jailbreaking Embodied {LLM} Agents in the Physical World},
  booktitle = {Proc. ICLR},
  year = {2025},
  url = {https://proceedings.iclr.cc/paper_files/paper/2025/hash/5b2fa23e4ef0f7ac6c4f01d7998e6237-Abstract-Conference.html},
}

@inproceedings{liu2025capo,
  author = {Liu, Jie and Zhou, Pan and Du, Yingjun and Tan, Ah-Hwee and Snoek, Cees G. M. and Sonke, Jan-Jakob and Gavves, Efstratios},
  title = {{CaPo}: Cooperative Plan Optimization for Efficient Embodied Multi-Agent Cooperation},
  booktitle = {Proc. ICLR},
  year = {2025},
  url = {https://proceedings.iclr.cc/paper_files/paper/2025/hash/b07091c16719ad3990e3d1ccee6641f1-Abstract-Conference.html},
}

@inproceedings{chang2025partnr,
  author = {Chang, Matthew and Chhablani, Gunjan and Clegg, Alexander and Cote, Mikael Dallaire and Desai, Ruta and Hlavac, Michal and Karashchuk, Vladimir and Krantz, Jacob and Mottaghi, Roozbeh and Parashar, Priyam and Patki, Siddharth and Prasad, Ishita and Puig, Xavier and Rai, Akshara and Ramrakhya, Ram and Tran, Daniel and Truong, Joanne and Turner, John and Undersander, Eric and Yang, Tsung-Yen},
  title = {{PARTNR}: A Benchmark for Planning and Reasoning in Embodied Multi-Agent Tasks},
  booktitle = {Proc. ICLR},
  year = {2025},
  url = {https://proceedings.iclr.cc/paper_files/paper/2025/hash/a3cf318fbeec1126da21e9185ae9908c-Abstract-Conference.html},
}

@article{lou2026agents,
  author = {Lou, Jiabin and Shi, Rongye and Wang, Haopeng and Yu, Ming-Ming and Wang, Yuanshuai and Wang, Qunbo and Wu, Wenjun},
  title = {{Agents Trainer}: Automatically Training Multi-Agent Reinforcement Learning Models for Drone Swarm Using Language Model-Based Agents},
  journal = {IEEE Trans. Autom. Sci. Eng.},
  volume = {23},
  pages = {8992--9006},
  year = {2026},
  
}

@article{li2026llmmrs,
  author = {Li, Peihan and An, Zijian and Abrar, Shams and Zhou, Lifeng},
  title = {Large Language Models for Multi-Robot Systems: A Survey},
  journal = {Auton. Robots},
  volume = {50},
  year = {2026},
  number = {3},
  note = {{Art. no.} 30},
}

@inproceedings{zhao2024expel,
  author = {Zhao, Andrew and Huang, Daniel and Xu, Quentin and Lin, Matthieu and Liu, Yong-Jin and Huang, Gao},
  title = {{ExpeL}: {LLM} Agents Are Experiential Learners},
  booktitle = {Proc. AAAI Conf. Artif. Intell.},
  volume = {38},
  number = {17},
  pages = {19632--19642},
  year = {2024},
  url = {https://ojs.aaai.org/index.php/AAAI/article/view/29936},
}

@misc{wang2026shaper,
  author = {Wang, Peidong and Ma, Zhiming and Chang, Ying and Luo, Xufang and Zhang, Yiqun and Wang, Zihan and Yang, Xiaocui and Feng, Shi and Yang, Yuqing and Li, Dongsheng},
  title = {Self-Evolving Embodied Agents via Skill-Harness Evolution},
  year = {2026},
  note = {arXiv:2608.11350},
  url = {https://arxiv.org/abs/2608.11350},
}

@inproceedings{iannoli2026mission,
  author = {Iannoli, Andrea and Gigli, Lorenzo and Sciullo, Luca and Trotta, Angelo and Di Felice, Marco},
  title = {Say the Mission, Execute the Swarm: Agent-Enhanced {LLM} Reasoning in the {Web-of-Drones}},
  year = {2026},
  pages = {139--148},
  booktitle = {Proc. IEEE Int. Symp. World Wireless, Mobile Multimedia Netw. (WoWMoM)},
}

@misc{lou2026aeris,
  author = {Lou, Jiabin and Wang, Haopeng and Liu, Xinyu and Zhang, Yu and Shi, Rongye and Wu, Wenjun},
  title = {{AERIS}: Aerial-Edge Role-Driven Intelligence at Runtime via Orchestrated Language-Model Swarm},
  year = {2026},
  note = {arXiv:2606.30151},
  url = {https://arxiv.org/abs/2606.30151},
}

@article{lou2025talker,
  author = {Lou, Jiabin and Shi, Rongye and Lin, Yuxin and Wang, Qunbo and Wu, Wenjun},
  journal = {IEEE Robot. Autom. Lett.},
  title = {{TALKER}: A Task-Activated Language Model Based Knowledge-Extension Reasoning System},
  year = {2025},
  volume = {10},
  number = {2},
  pages = {1026--1033},
}

@misc{lin2026skillmas,
  author = {Lin, Hehai and Yang, Qi and Qin, Chengwei},
  title = {{Skill-MAS}: Evolving Meta-Skill for Automatic Multi-Agent Systems},
  year = {2026},
  url = {https://arxiv.org/abs/2606.18837},
  note = {arXiv:2606.18837},
}

@inproceedings{yao2023react,
  author = {Yao, Shunyu and Zhao, Jeffrey and Yu, Dian and Du, Nan and Shafran, Izhak and Narasimhan, Karthik and Cao, Yuan},
  title = {{ReAct}: Synergizing Reasoning and Acting in Language Models},
  booktitle = {Proc. ICLR},
  year = {2023},
  url = {https://openreview.net/forum?id=WE_vluYUL-X},
}

@inproceedings{liang2023codepolicies,
  author = {Liang, Jacky and Huang, Wenlong and Xia, Fei and Xu, Peng and Hausman, Karol and Ichter, Brian and Florence, Pete and Zeng, Andy},
  title = {Code as Policies: Language Model Programs for Embodied Control},
  booktitle = {Proc. IEEE Int. Conf. Robot. Autom. (ICRA)},
  pages = {9493--9500},
  year = {2023},
  url = {https://code-as-policies.github.io/},
}

@inproceedings{huang2025rekep,
  author = {Huang, Wenlong and Wang, Chen and Li, Yunzhu and Zhang, Ruohan and Fei-Fei, Li},
  title = {{ReKep}: Spatio-Temporal Reasoning of Relational Keypoint Constraints for Robotic Manipulation},
  booktitle = {Proc. Conf. Robot Learn. (CoRL)},
  series = {PMLR},
  volume = {270},
  pages = {4573--4602},
  year = {2025},
  url = {https://proceedings.mlr.press/v270/huang25g.html},
}

@inproceedings{xu2025amem,
  author = {Xu, Wujiang and Liang, Zujie and Mei, Kai and Gao, Hang and Tan, Juntao and Zhang, Yongfeng},
  title = {{A-Mem}: Agentic Memory for {LLM} Agents},
  booktitle = {Adv. Neural Inf. Process. Syst.},
  volume = {38},
  year = {2025},
  url = {https://proceedings.neurips.cc/paper_files/paper/2025/hash/19909c36f51abc4856b4560aff3d36d6-Abstract-Conference.html},
  pages = {20004--20031},
}

@misc{li2026jitrl,
  author = {Li, Yibo and Lin, Zijie and Deng, Ailin and Zhang, Xuan and He, Yufei and Ji, Shuo and Cao, Tri and Hooi, Bryan},
  title = {Just-In-Time Reinforcement Learning: Continual Learning in {LLM} Agents Without Gradient Updates},
  year = {2026},
  url = {https://arxiv.org/abs/2601.18510},
  note = {arXiv:2601.18510},
}

@inproceedings{kannan2024smartllm,
  author = {Kannan, Shyam Sundar and Venkatesh, Vishnunandan L. N. and Min, Byung-Cheol},
  title = {{SMART-LLM}: Smart Multi-Agent Robot Task Planning using Large Language Models},
  booktitle = {Proc. IEEE/RSJ Int. Conf. Intell. Robots Syst. (IROS)},
  pages = {12140--12147},
  year = {2024},
  url = {https://ieeexplore.ieee.org/document/10802322},
}

@inproceedings{zhang2024coela,
  author = {Zhang, Hongxin and Du, Weihua and Shan, Jiaming and Zhou, Qinhong and Du, Yilun and Tenenbaum, Joshua B. and Shu, Tianmin and Gan, Chuang},
  title = {Building Cooperative Embodied Agents Modularly with Large Language Models},
  booktitle = {Proc. ICLR},
  year = {2024},
  url = {https://proceedings.iclr.cc/paper_files/paper/2024/hash/54b8b4e0b4ba4aad112e84f32e3b5dbb-Abstract-Conference.html},
}

@inproceedings{mandi2024roco,
  author = {Mandi, Zhao and Jain, Shreeya and Song, Shuran},
  title = {{RoCo}: Dialectic Multi-Robot Collaboration with Large Language Models},
  booktitle = {Proc. IEEE Int. Conf. Robot. Autom. (ICRA)},
  year = {2024},
  url = {https://ieeexplore.ieee.org/document/10610855/},
  pages = {286--299},
}

@misc{deepseekai2026deepseekv4,
  author = {{DeepSeek-AI}},
  title = {{DeepSeek-V4}: Towards Highly Efficient Million-Token Context Intelligence},
  year = {2026},
  url = {https://arxiv.org/abs/2606.19348},
  note = {arXiv:2606.19348},
}

\end{document}